\documentclass[letterpaper]{article} % DO NOT CHANGE THIS
\usepackage{aaai2026}  % DO NOT CHANGE THIS
\usepackage{times}  % DO NOT CHANGE THIS
\usepackage{helvet}  % DO NOT CHANGE THIS
\usepackage{courier}  % DO NOT CHANGE THIS
\usepackage[hyphens]{url}  % DO NOT CHANGE THIS
\usepackage{graphicx} % DO NOT CHANGE THIS
\usepackage{natbib}  % DO NOT CHANGE THIS AND DO NOT ADD ANY OPTIONS TO IT
\usepackage{caption} % DO NOT CHANGE THIS AND DO NOT ADD ANY OPTIONS TO IT
\usepackage{algorithm}
\usepackage{algorithmic}
\usepackage{booktabs}
\usepackage{colortbl}
\usepackage{multirow}
\usepackage{amssymb}
\usepackage{amsmath}
\usepackage{newfloat}
\usepackage{listings}

\DeclareCaptionStyle{ruled}{labelfont=normalfont,labelsep=colon,strut=off} % DO NOT CHANGE THIS
\floatstyle{ruled}
\newfloat{listing}{tb}{lst}{}
\floatname{listing}{Listing}
\title{PCSR: Pseudo-label Consistency-Guided Sample Refinement for Noisy Correspondence Learning}
\author{
    Zhuoyao Liu\textsuperscript{\rm 1},
    Yang Liu\textsuperscript{\rm 1},
    Wentao Feng\textsuperscript{\rm 1},
    Shudong Huang\textsuperscript{\rm 1,2}\thanks{Corresponding author.},\\
}

\affiliations{
    \textsuperscript{\rm 1}College of Computer Science, Sichuan University, China\\
    \textsuperscript{\rm 2}Engineering Research Center of Machine Learning and Industry Intelligence, China\\
    liuzhuoyao@stu.scu.edu.cn
}

\usepackage{bibentry}
\begin{document}

\maketitle

\begin{abstract}
Cross-modal retrieval aims to align different modalities via semantic similarity. 
However, existing methods often assume that image-text pairs are perfectly aligned, overlooking \textit{Noisy Correspondences} in real data. 
These misaligned pairs misguide similarity learning and degrade retrieval performance. 
Previous methods often rely on coarse-grained categorizations that simply divide data into clean and noisy samples, overlooking the intrinsic diversity within noisy instances. 
Moreover, they typically apply uniform training strategies regardless of sample characteristics, resulting in suboptimal sample utilization for model optimization.
To address the above challenges, we introduce a novel framework, called \emph{Pseudo-label Consistency-Guided Sample Refinement (PCSR)}, which enhances correspondence reliability by explicitly dividing samples based on pseudo-label consistency.
Specifically, we first employ a confidence-based estimation to distinguish clean and noisy pairs, then refine the noisy pairs via pseudo-label consistency to uncover structurally distinct subsets.
We further propose a \emph{Pseudo-label Consistency Score} (PCS) to quantify prediction stability, enabling the separation of ambiguous and refinable samples within noisy pairs. 
Accordingly, we adopt \emph{Adaptive Pair Optimization} (APO), where ambiguous samples are optimized with robust loss functions and refinable ones are enhanced via text replacement during training.
Extensive experiments on CC152K, MS-COCO and Flickr30K validate the effectiveness of our method in improving retrieval robustness under noisy supervision.
\end{abstract}


\section{Introduction}

\begin{figure}[t]
\centering
\includegraphics[width=1\columnwidth]{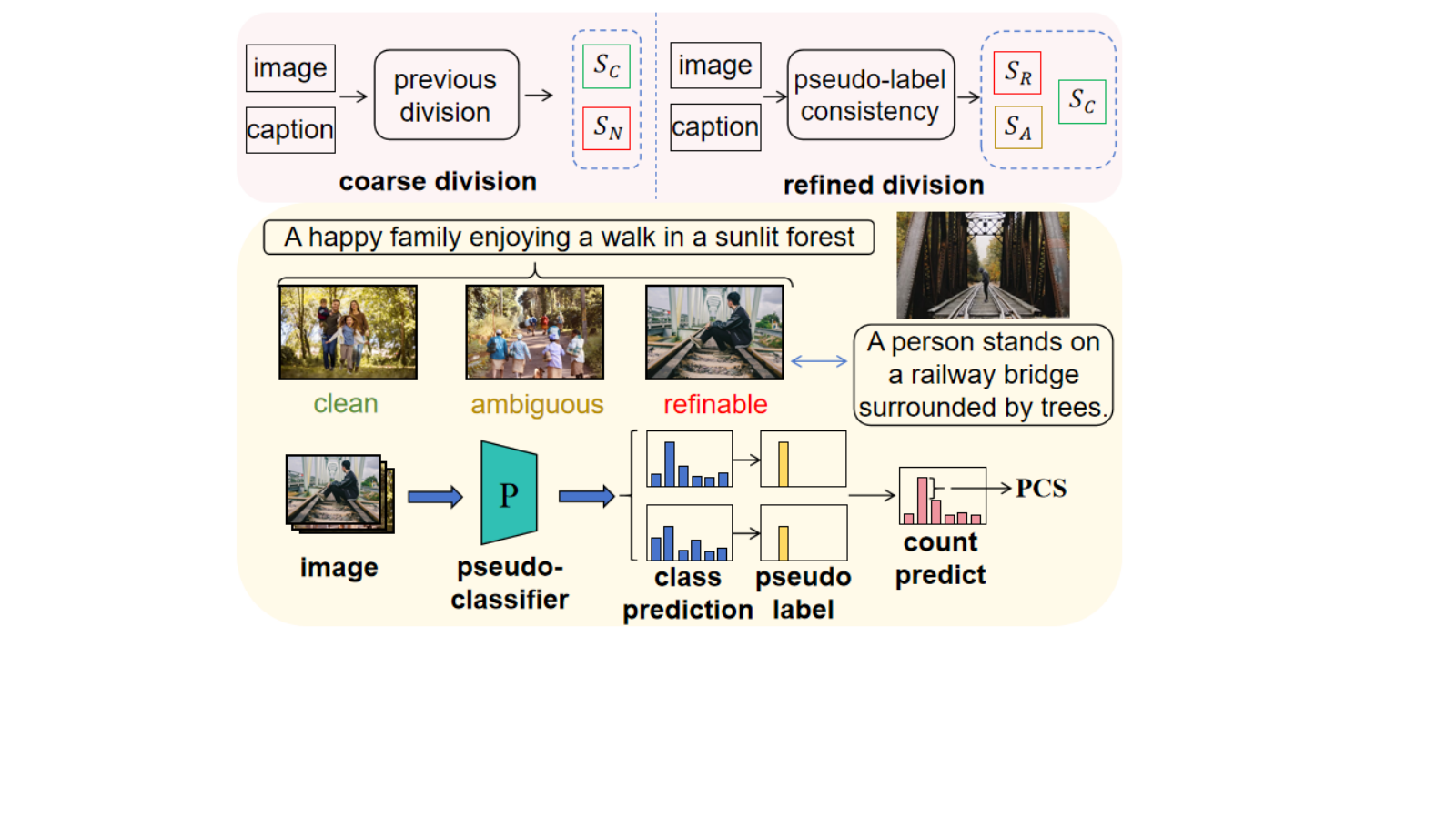} % Reduce the figure size so that it is slightly narrower than the column. Don't use precise values for figure width.This setup will avoid overfull boxes.
\caption{(Top) Difference between previous coarse division and our refined division method. (Bottom) The conceptual diagram of our proposed PCSR method. Specifically, we first obtain the pseudo-label of each sample using a pseudo-classifier, and then calculate PCS value by measuring the concentration of the pseudo-label prediction. Thereby distinguishing samples with consistent image predictions.}
\label{fig1}
\end{figure}

Cross-modal retrieval \cite{anderson2018bottom,faghri2017vse++}, a fundamental task in multimodal learning, aims to bridge different modalities within large-scale multimedia data. Existing methods explore various strategies to bridge the semantic gap between modalities \cite{feng2023rono,liu2020graph,zhang2020context,liu2025asymmetric}, but their effectiveness relies heavily on the assumption of accurately aligned image-text pairs. In practice, such precise alignment requires costly manual annotation, which is both challenging and error-prone. This results in the problem of noisy correspondence, different from traditional noisy labels \cite{jiang2020beyond,bai2021understanding,iscen2022learning,yan2023adaptive} that refer to incorrect category annotations. Noisy correspondence specifically refers to semantic misalignment between paired modalities, which consequently degrades the performance of existing cross-modal retrieval methods \cite{diao2021similarity,li2019visual,lee2018stacked}.

In recent years, research addressing the problem of noisy correspondence remains limited. Earlier methods \cite{qin2022deep,yang2023bicro} typically divide the training data into noisy and clean subsets, and then employ strategies such as dynamic weight adjustment to mitigate the impact of incorrect supervision or introduce effective supervisory signals to enhance model optimization. However, these methods only consider a binary categorization and fail to account for the possibility that some pseudo-positive pairs within the noisy subset may still exhibit partial semantic alignment and thus can still provide informative supervision while training. To address this, \cite{ma2024cross} introduces a finer-grained classification that partitions samples into clean, vague, and noisy categories. However, such a division remains ambiguous, lacking quantifiable criteria such as prediction stability or semantic consistency to justify the classification. This approach merely refines existing confidence-based or loss-based frameworks without introducing fundamentally new classification principles. Moreover, in the subsequent training phase, coarse division approaches tend to apply uniform training strategies across all noisy pairs, which leads to two major drawbacks: stable noisy samples are underutilized, and ambiguous samples introduce incorrect supervision, thereby affecting the convergence of the model. The training strategies proposed after classification often only involve parameter adjustments compared to the original methods. Consequently, although these methods alleviate noisy supervision through finer categorization, the refinement exerts limited impact on the overall training strategy and offers restricted performance gains.

As illustrated in Figure 1, we propose a refined division strategy that categorizes samples into three subsets: clean, ambiguous, and refinable. We argue that there are intrinsic differences within the noisy subset: some are reliably identified as mismatches due to accurate model understanding and are thus termed as refinable pairs, while others are classified as noisy due to semantic ambiguity and are referred to as ambiguous pairs. For refinable pairs, which the model understands confidently, we perform text refinement to enhance their utility. For ambiguous pairs, where the model’s predictions are uncertain, we adopt a more robust training strategy. This approach enables us to fully exploit the distinct training value inherent in each category.

To address these challenges, we propose a novel method named \textbf{\textit{Pseudo-label Consistency-Guided Sample Refinement}} (PCSR), which assesses correspondence reliability using confidence scores and pseudo-label consistency, thereby improving the robustness of cross-modal retrieval by enabling flexible sample division and targeted training strategies. Specifically, PCSR first estimates the confidence of each image-text pair based on the distribution of matching losses, using Gaussian Mixture Model (GMM) to distinguish clean pairs and noisy pairs. Considering some noisy pairs contain useful information that can be exploited through training, as illustrated in Figure 1, we further introduce a \emph{Pseudo-label Consistency Score} (PCS) in our method to analyze the prediction stability of images within noisy pairs. By evaluating the consistency across multiple pseudo-label predictions, we separate noisy pairs into two different subsets: Ambiguous pairs that suffer from unstable pseudo-classifier  predictions, implying limited training value. Conversely, refinable pairs exhibit stable predictions, suggesting that their images remain valuable for learning despite misaligned textual annotations. Building on this division, PCSR adopts \emph{Adaptive Pair Optimization} (APO). Whereas clean pairs are trained using standard triplet loss and cross-entropy loss. For ambiguous pairs, to mitigate the impact of unstable supervision, we adopt a more robust Generalized Cross-Entropy (GCE) loss for training. For refinable pairs, benefiting from reliable image pseudo-prediction, we undergo a caption rematch strategy, where their original captions are replaced by captions retrieved from clean pairs based on pseudo-label similarity. This allows the model to receive more reliable supervision signals.

Our main contributions are summarized as follows:

\begin{itemize}

\item Inspired by the observation that noisy pairs exhibit intrinsic differences, we propose a novel \emph{Pseudo-label Consistency-Guided Sample Refinement} (PCSR) framework, which explicitly models correspondence reliability through a refined sample division strategy.

\item We propose an \emph{Adaptive Pair Optimization} (APO) strategy, introducing distinct optimization processes tailored to clean, ambiguous, and refinable pairs, thereby fully exploiting the particular training value of each subset.

\item We conduct extensive experiments on MS-COCO, Flickr30K and CC152K. Experimental results show that PCSR effectively obtains reliable supervisory signals from each identified category and significantly outperforms state-of-the-art methods under noisy conditions.

\end{itemize}

\section{Related Work}

\begin{figure*}[t]
\centering
\includegraphics[width=1.0\textwidth]{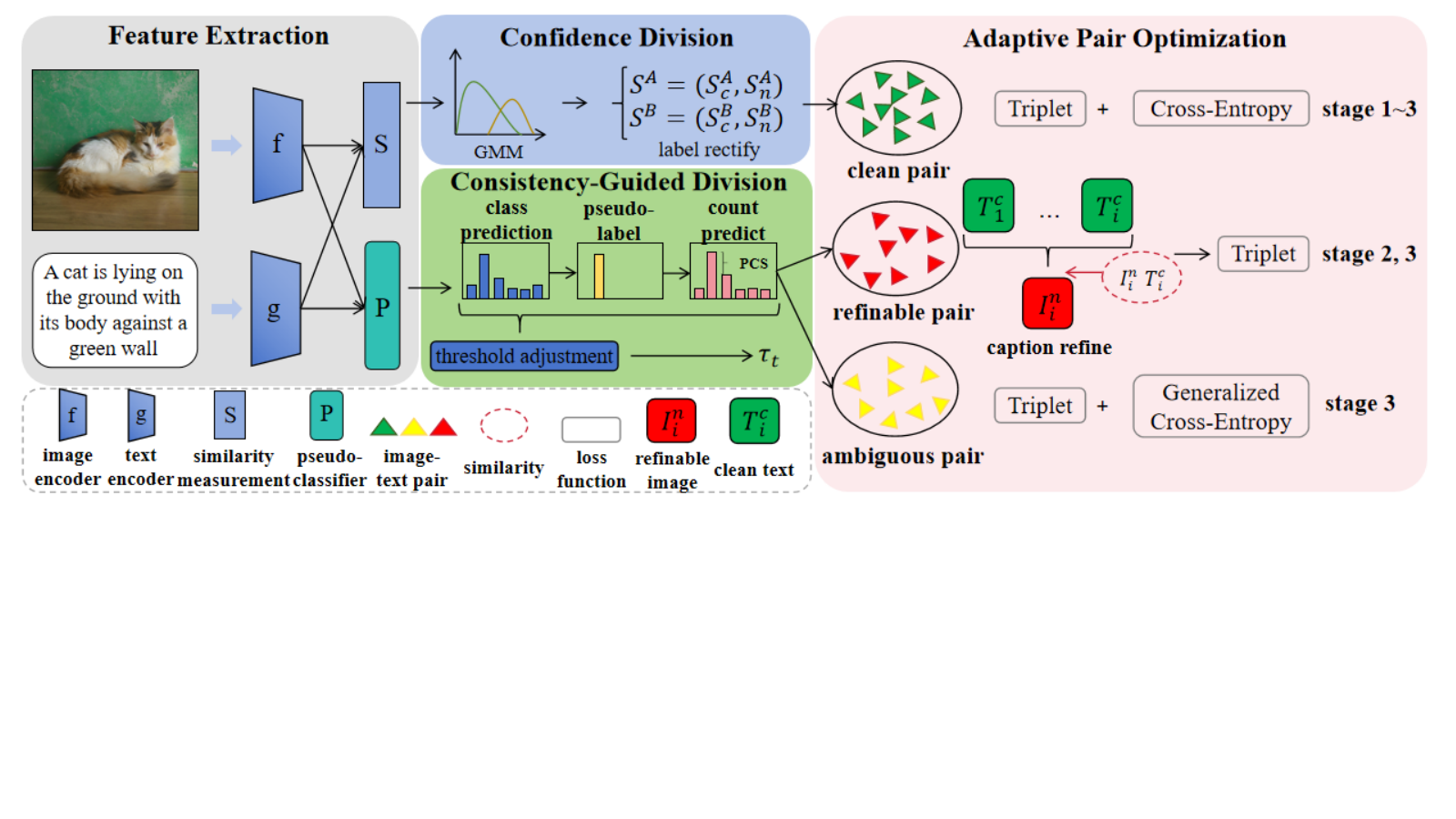}
\caption{Overview of the proposed method. We compute the Pseudo-label Consistency Score (PCS) through repeated pseudo-label predictions, further dividing the noisy pairs into ambiguous pairs and refinable pairs. For clean pairs, we employ a combination of triplet loss and cross-entropy loss for optimization. For ambiguous pairs, we utilize triplet loss alongside generalized cross-entropy (GCE) loss to enhance robustness against label uncertainty. For refinable pairs, given their consistent pseudo-label predictions across iterations, we discard their original texts and replace them with captions selected from a subset of clean pairs that are most semantically similar to the noisy image’s pseudo-label, thereby enabling more reliable training.
}
\label{fig2}
\end{figure*}

% \subsection*{Cross-Modal Matching}
\noindent \textbf{Cross-Modal Matching}
Cross-modal matching \cite{wang2020cross,yang2022dual} is a core task in information retrieval \cite{zhao2024seer} and multi-modal learning \cite{chuang2022robust}, aiming to retrieve relevant content across different modalities using one modality as a query. Among these tasks, image-text retrieval is the most representative and is typically divided into two mainstream paradigms: coarse-grained and fine-grained alignment. Coarse-grained methods \cite{Liu_2025_ICCV,chen2021learning,li2019visual} map entire images and texts into a common embedding space using separate networks and compute their overall similarity primarily based on cosine distance. Fine-grained approaches \cite{lee2018stacked,zhang2022show,zhang2022negative,pan2023fine,fu2024linguistic} emphasize cross-modal interaction and align local regions across modalities, subsequently combining local matching results to derive the final similarity score. Despite their effectiveness, these methods generally rely on large-scale, accurately aligned training pairs and largely ignore the critical issue of noisy correspondence in practical scenarios. Therefore, exploring how to effectively learn cross-modal retrieval in the noisy correspondence relationships is of great significance, but this challenging issue has rarely been touched upon in previous studies.

% \subsection*{Noisy Correspondence Learning}
\noindent\textbf{Noisy Correspondence Learning}
Noisy correspondence refers to the widespread phenomenon where semantically misaligned data pairs are erroneously regarded as matched pairs during training. NCR \cite{huang2021learning} represents the first pioneering work specifically dedicated to addressing this problem. NCR enhances the robustness of image-text matching models by using a Gaussian Mixture Model (GMM) to adaptively refine soft correspondence labels during training. Subsequently, BiCro \cite{yang2023bicro} uses Beta Mixture Model (BMM) to refine binary noisy labels into soft labels, assuming similar images share similar texts and vice versa. MSCN \cite{han2023noisy} uses clean data to train a meta-network for generating reliable soft labels to correct main network predictions. Other recent methods \cite{qin2022deep,li2024nac} enhance the loss function using dynamic weighting or bi-directional similarity. However, the existing methods mentioned above largely ignored structural differences among noisy pairs, training all noisy pairs using the same homogeneous method inherently limits the exploitation of their varied optimization potential. 
In contrast, we introduce a refined division strategy that effectively classifies noisy samples, allowing more precise distinction based on their inherent differences. Meanwhile, our method applies tailored training strategies to each category, enabling better utilization of noisy pairs based on their unique optimization potential.

% \subsection*{Pseudo-Labeling}
\noindent\textbf{Pseudo-Labeling}
Pseudo-labeling was initially introduced to deep learning by \cite{lee2013pseudo}. It was first used in the field of computer vision and later widely adopted in semi-supervised learning and unsupervised learning. Semi-supervised methods \cite{chen2024fast,chen2024multi,heidari2024reinforcement} often combined with consistency regularization, label propagation, and multi-model collaboration. These methods enhance pseudo-label quality and robustness by enforcing consistency, leveraging weak supervision, and exploiting model complementarity. Pseudo-labels are also used in unsupervised learning fields like self-supervised learning \cite{zhang2024improving} and knowledge distillation \cite{10377550,WEN202125}, where models generate supervisory signals to guide representation learning or knowledge transfer. Further on, pseudo-labeling inherently tolerates label noise, as many of its techniques \cite{8417973,Wang_Han_Liu_Niu_Yang_Gong_2021} are inspired by or integrate noise-robust learning strategies to handle imperfect labels during training.
Our method, building upon PC2 \cite{duan2024pc2}, applies pseudo-labeling to cross-modal retrieval and achieves significant performance improvements.

\section{Method}

\subsection*{Problem Overview}

Given a query, cross-modal retrieval aims to identify semantically relevant instances from a dataset of a different modality. Without loss of generality, we take image-text retrieval as an example to discuss the problem of noisy correspondence in cross-modal retrieval. As illustrated in \textbf{Figure 2}, our proposed method comprises four key components. The feature extraction and confidence division parts are derived from the baseline strategy introduced in NCR \cite{huang2021learning}. Specifically, given the image-text dataset, we use an image encoder $f(\cdot)$ and a text encoder $g(\cdot)$ to compute the feature embedding into a shared embedding space, then the similarity of image-text pairs is calculated within $S(f(I),g(T))$, where $I$ and $T$ represent images and texts, and $S$ indicates the similarity function. Following, confidence division derives a probability distribution using a Gaussian Mixture Model (GMM) to divide the samples into clean samples $(y_i=1)$ and noisy samples $(y_i=0)$. To fully exploit the distinct training value of all samples, we propose a refined division method.

To better assess the reliability of image-text pairs under noisy correspondence, we adopt a \emph{Consistency-Guided Division} (CGD) strategy. Inspired by semi-supervised learning, we propose a pseudo-classifier $C (\cdot)$ to predict pseudo categories of each image and text, based on its prediction consistency, we further divide noisy samples into refinable and ambiguous samples. Following this, we propose an \emph{Adaptive Pair Optimization} (APO) strategy that performs targeted training based on the distinctive properties of the data.

% For clarity, we first introduce the mathematical symbols used. Formally, we use $\mathcal{D} = \left\{ (I_i, T_i, y_i) \right\}_{i=1}^N$ 
% to represent an image-text matching dataset. $I_i$ is the $i$-th image in a training batch, $T_i$ is the $i$-th text in a training batch. Binary label $y_i \in \{0, 1\}$ indicates the corresponding label given by confidence division shown in figure 2, $N$ represents the size of our training set. 

\subsection{Consistency-Guided Division}

% The core idea is to leverage a pseudo-classifier to repeatedly predict category labels for each modality and track their prediction consistency over time. Intuitively,

If the classifier consistently assigns the same category to an image across epochs, it suggests strong semantic confidence, whereas fluctuating predictions indicate ambiguity. Based on this principle, we propose a refined division strategy named \emph{Consistency-Guided Division} (CGD). By training a pseudo classifier within clean-only epochs, we define a Pseudo-label Consistency Score (PCS) to measure label stability and dynamically adjust a threshold $\tau$ to separate ambiguous and refinable samples. This module enables the model to selectively refine informative noisy pairs while avoiding misleading supervision from unstable ones. Thus, the model's understanding of image-text pairs can be assessed by measuring the consistency of its prediction.

\paragraph{Pseudo-Classifier Training on clean data.} Consistency-Guided Division relies heavily on category predictions from a pseudo-classifier, making it a critical component of the overall framework. Existing studies \cite{duan2024pc2} applied pseudo-labeling in cross-modal retrieval, and proposed a method for training a pseudo-classifier. Using only the clean data, we train the pseudo-classifier by utilizing predicted category $c$ for each image and the pseudo-label $l \in \{1, \cdots, K\}$ of its corresponding caption. $K$ represents the number of categories predicted by the Pseudo-Classifier.

Specifically, after the confidence division step, the dataset is divided into clean pairs and noisy pairs. Given a batch of clean pairs ${(I_i^c, T_i^c)}_{i=1}^B$, following the encoding process, the pseudo-classifier is subsequently employed to produce pseudo-label predictions for image ${p}_i^{,c} = C(f(I_i^{,c}))$ and text ${q}_i^{,c} = C(g(T_i^{,c}))$. We obtain the pseudo-label for the text by selecting the class with the highest predicted probability $\hat{q}_i^c = \arg\max(q_i^c)$, then optimize the pseudo-classifier using the standard cross-entropy loss with pseudo-label as the target:

\begin{equation}
\mathcal{L}_{\text{CE}} = \frac{1}{B} \sum_{i=1}^{B} \left( - \sum_{k=1}^{K} \hat{q}_{i,k}^c \log(p_{i,k}^c) \right)
\end{equation}
To ensure diverse pseudo-label assignments, we regularize the classifier $\mathbf{C}$ with an entropy loss that encourages uniform distribution over the label space:

\begin{equation}
\mathcal{L}_{\text{En}} = -\frac{1}{B} \sum_{i=1}^{B} p_i^{\,c} \cdot \log\left( \frac{1}{B} \sum_{j=1}^{B} p_j^{\,c} \right)
\end{equation}

\paragraph{Pseudo-label Consistency Score.}
After obtaining the pseudo-predictions of images and texts, it is necessary to define a metric that quantifies the model's level of understanding for each sample. We posit that the more consistent the model's pseudo-label predictions for an image across multiple training iterations, the greater its semantic confidence in understanding the image. We first obtain the pseudo-label of the image $\hat{p}_i^c = \arg\max(p_i^c)$. In each training epoch, we count the occurrences of each label category as $n_1, n_2, \ldots, n_K$, where $n_K$ denotes the number of times class $K$ is predicted. We define the \emph{Pseudo-label Consistency Score} (PCS) as a metric to measure the prediction stability of the pseudo-classifier over multiple forward passes. Specifically, assume that $Q_i = [n_1, n_2, \ldots, n_K]$ represents the cumulative pseudo-label prediction count of each class for the image $I_i$ over multiple training epochs. The Pseudo-label Consistency Score (PCS) for $I_i$ is defined as:
\begin{equation}
\mathrm{PCS}_i = n_{\max} - n_{\mathrm{second}},
\end{equation}
where $n_{\max} = \max(Q_i)$ denotes the number of times the most frequently predicted class appears, and $n_{\mathrm{second}} = \max(Q_i \setminus \{n_{\max}\})$ corresponds to the count of the second most frequent prediction. A larger $\mathrm{PCS}_i$ implies higher consistency in pseudo-label predictions, reflecting stronger semantic confidence from the model regarding $I_i$.

\paragraph{Dynamic Threshold Adjustment.} To ensure a stable and effective separation of ambiguous and refinable samples, we design a dynamic threshold adjustment mechanism that adaptively updates the threshold $\tau$ during training. Among noisy samples with low confidence, we define  \emph{refinable} samples $S_R$ as follows:
\[
\mathcal{S}_{R} = \left\{ (I_i,T_i) \,\middle|\, y_i=0,\; \text{PCS}_i \geq \tau \right\}
\]
and \emph{ambiguous} samples $S_A$ as follows:
\[
\mathcal{S}_{A} = \left\{ (I_i,T_i) \,\middle|\, y_i=0,\; \text{PCS}_i < \tau \right\}
\]
This mechanism is based on the target utilization rate $\lambda_{\text{target}}$, which gradually increases as training progresses:

\begin{equation}
\lambda_{\text{target}} = \lambda_{\min} + (\lambda_{\max} - \lambda_{\min}) \cdot \frac{t}{T}
\end{equation}
where $t$ is the current training iteration and $T$ is the total number of training iterations. This encourages the model to utilize more samples progressively. Following, we compute the intermediate target threshold $\tau_{\text{target}}$ based on the discrepancy between current utilization $\lambda$ and the expected target:

\begin{equation}
\tau_{\text{target}} = \tau - k \cdot (\lambda_{\text{target}} - \lambda)
\end{equation}
Here, $k$ is a sensitivity coefficient that controls the adjustment strength. If the actual utilization is lower than expected, $\tau$ will be reduced to include more samples, and vice versa. In this work, we update the threshold of current epoch $\tau_t$ using an Exponential Moving Average (EMA) to ensure smoothness:

\begin{equation}
\tau_t = (1 - \beta) \cdot \tau_{t-1} + \beta \cdot \tau_{\text{target}}
\end{equation}
where $\beta$ is the parameter controlling the update rate.

\subsection{Adaptive Pair Optimization}

\begin{table*}[t]
\centering
\setlength{\tabcolsep}{5pt}
\scriptsize
\renewcommand{\arraystretch}{0.9}  % 缩小行间距
\begin{tabular}{c|l|ccc|ccc|c||ccc|ccc|c}
\toprule
\multicolumn{2}{c|}{} & \multicolumn{7}{c||}{\textbf{Flickr30K}} & \multicolumn{7}{c}{\textbf{MSCOCO 1K}} \\
\cmidrule(lr){3-9} \cmidrule(lr){10-16}
\multirow{2}{*}{\centering \textbf{Noise}} & \multirow{2}{*}{\centering \textbf{Methods}} 
& \multicolumn{3}{c|}{Image$\rightarrow$Text} 
& \multicolumn{3}{c|}{Text$\rightarrow$Image} 
& \multirow{2}{*}{\centering Rsum} 
& \multicolumn{3}{c|}{Image$\rightarrow$Text} 
& \multicolumn{3}{c|}{Text$\rightarrow$Image} 
& \multirow{2}{*}{\centering Rsum} \\
& & R@1 & R@5 & R@10 & R@1 & R@5 & R@10 & & R@1 & R@5 & R@10 & R@1 & R@5 & R@10 & \\

\midrule
\multirow{9}{*}{20\%}
& SCAN (ECCV’18) & 59.1 & 83.4 & 90.4 & 36.6 & 67.0 & 77.5 & 414.0 & 66.2 & 91.0 & 96.4 & 45.0 & 80.2 & 89.3 & 468.1 \\
& SGR (AAAI’21) & 61.2 & 84.3 & 91.5 & 44.5 & 72.1 & 80.2 & 433.8 & 49.1 & 83.8 & 92.7 & 42.5 & 77.7 & 88.2 & 434.0 \\
& NCR* (NIPS’21) & 73.5 & 93.2 & 96.6 & 56.9 & 82.4 & 88.5 & 491.1 & 76.6 & 95.6 & 98.2 & 60.8 & 88.8 & 95.0 & 515.0 \\
& DECL* (ACMMM’22) & 77.5 & 93.8 & 97.0 & 56.1 & 81.8 & 88.5 & 494.7 & 77.5 & 95.9 & 98.4 & 61.7 & 89.3 & 95.4 & 518.2 \\
& MSCN* (CVPR’23) & 77.4 & 94.9 & 97.6 & 59.6 & 83.2 & 89.2 & 501.9 & 78.1 & \underline{97.2} & 98.8 & 64.3 & \underline{90.4} & \textbf{95.8} & 524.6 \\
& BiCro* (CVPR’23) & 78.1 & 94.4 & 97.5 & 60.4 & \textbf{84.4} & 89.9 & 504.7 & 78.8 & 96.1 & 98.6 & 63.7 & 90.3 & \underline{95.7} & 523.2 \\
& RCL* (TPAMI’23) & 75.9 & 94.5 & 97.3 & 57.9 & 82.6 & 88.6 & 496.8 & 78.9 & 96.0 & 98.4 & 62.8 & 89.9 & 95.4 & 521.4 \\
& CREAM* (TIP’24) & 77.4 & 95.0 & 97.5 & 58.7 & 84.1 & 89.8 & 502.3 & 78.9 & 96.3 & \underline{98.7} & 63.3 & \textbf{90.5} & 95.3 & 523.0 \\
& L2RM* (CVPR’24) & 77.9 & \underline{95.2} & \textbf{97.8} & 59.8 & 83.6 & 89.5 & 503.8 & \underline{80.2} & 96.3 & 98.5 & \textbf{64.2} & 90.1 & 95.4 & \underline{524.7} \\
& \textbf{PCSR*} & \textbf{78.7} & \underline{95.1} & \underline{97.9} & \textbf{60.9} & \underline{83.7} & \textbf{89.4} & \textbf{505.7} & \textbf{80.5} & \textbf{97.3} & \textbf{98.9} & \underline{63.8} & 89.7 & 95.1 & \textbf{525.4} \\
\midrule
\multirow{9}{*}{40\%}
& SCAN (ECCV’18) & 29.9 & 60.5 & 72.5 & 16.4 & 38.5 & 48.6 & 266.4 & 30.1 & 65.2 & 79.2 & 18.9 & 51.1 & 69.9 & 314.4 \\
& SGR (AAAI’21) & 47.2 & 76.4 & 83.2 & 34.5 & 60.3 & 70.5 & 372.1 & 43.9 & 78.3 & 89.3 & 37.0 & 72.8 & 85.1 & 406.4 \\
& NCR* (NIPS’21) & 75.3 & 92.1 & 95.2 & 56.2 & 80.6 & 87.4 & 486.8 & 76.5 & 95.0 & 98.2 & 60.7 & 88.5 & 95.0 & 513.9 \\
& DECL* (ACMMM’22) & 72.7 & 92.3 & 95.4 & 53.4 & 79.4 & 86.4 & 479.6 & 75.6 & 95.5 & 98.3 & 59.5 & 88.3 & 95.4 & 512.0 \\
& MSCN* (CVPR’23) & 74.4 & 94.4 & 96.9 & 57.2 & 81.7 & 87.6 & 492.2 & 74.8 & 94.9 & 98.0 & 60.3 & 88.5 & 94.3 & 510.9 \\
& BiCro* (CVPR’23) & 74.6 & 92.7 & 96.2 & 55.5 & 81.0 & 87.4 & 487.5 & 77.0 & \underline{95.9} & 98.3 & 61.1 & 89.2 & 94.9 & 517.1 \\
& RCL* (TPAMI’23) & 72.7 & 92.7 & 96.1 & 54.8 & 80.0 & 87.1 & 483.4 & 77.0 & 95.5 & \underline{98.4} & 61.2 & \textbf{90.2} & 95.1 & 515.3 \\
& CREAM* (TIP’24) & \underline{76.3} & \underline{93.3} & \underline{97.1} & \underline{57.0} & \textbf{82.6} & \underline{88.7} & \underline{495.0} & 76.5 & 95.0 & 98.2 & 61.7 & \underline{89.4} & \textbf{95.6} & 516.8 \\
& L2RM* (CVPR’24) & 75.8 & 93.2 & 96.9 & 56.3 & 81.0 & 87.3 & 490.5 & \underline{77.5} & 95.8 & \textbf{98.5} & \underline{62.0} & 89.1 & 94.9 & \underline{517.7} \\
& \textbf{PCSR*} & \textbf{76.6} & \textbf{94.5} & \textbf{97.3} & \textbf{57.4} & \underline{82.4} & \textbf{88.9} & \textbf{497.1} & \textbf{77.7} & \textbf{96.0} & 98.1 & \textbf{63.1} & \underline{89.4} & \underline{95.5} & \textbf{519.8} \\
\midrule
\multirow{9}{*}{60\%}
& SCAN (ECCV’18) & 16.9 & 39.3 & 53.9 & 2.8 & 7.4 & 11.4 & 131.7 & 27.8 & 59.8 & 74.8 & 16.8 & 47.8 & 66.4 & 293.4 \\
& SGR (AAAI’21) & 28.7 & 58.0 & 71.0 & 23.8 & 49.5 & 60.7 & 291.7 & 37.6 & 73.3 & 86.3 & 33.8 & 68.6 & 81.7 & 381.3 \\
& NCR* (NIPS’21) & 68.7 & 89.9 & 95.5 & 52.0 & 77.6 & 84.9 & 468.6 & 72.7 & 94.0 & 97.6 & 57.9 & 87.0 & 94.1 & 503.3 \\
& DECL* (ACMMM’22) & 65.2 & 88.4 & 94.0 & 46.8 & 74.0 & 82.2 & 450.6 & 73.0 & 94.2 & 97.1 & 57.0 & 86.8 & 94.0 & 502.5 \\
& MSCN* (CVPR’23) & 70.4 & 91.0 & 94.9 & 53.4 & 77.8 & 84.1 & 471.6 & 74.4 & 95.1 & 97.9 & 59.2 & 87.1 & 92.8 & 506.5 \\
& BiCro* (CVPR’23) & 67.6 & 90.0 & 94.4 & 51.2 & 77.6 & 84.7 & 466.3 & 73.9 & 94.4 & 97.8 & 58.3 & 87.2 & 93.5 & 505.5 \\
& RCL* (TPAMI’23) & 67.7 & 89.1 & 93.6 & 48.0 & 74.9 & 83.3 & 456.6 & 74.0 & 94.3 & 97.6 & 57.6 & 86.4 & 93.5 & 503.3 \\
& CREAM* (TIP’24) & \underline{70.6} & \underline{91.2} & \textbf{96.1} & \underline{53.3} & \underline{79.2} & \textbf{87.0} & \underline{477.4} & 74.7 & \underline{94.7} & \textbf{98.0} & \underline{59.7} & \textbf{88.0} & \textbf{94.6} & \underline{509.9} \\
& L2RM* (CVPR’24) & 70.0 & 90.8 & 95.4 & 51.3 & 76.4 & 83.7 & 467.6 & \underline{75.4} & \underline{94.7} & \underline{97.9} & 59.2 & 87.4 & 93.6 & 508.4 \\
& \textbf{PCSR*} & \textbf{72.8} & \textbf{91.9} & \underline{95.8} & \textbf{54.8} & \textbf{79.7} & \underline{86.5} & \textbf{480.5} & \textbf{76.3} & \textbf{94.8} & \underline{97.9} & \textbf{61.7} & \underline{87.5} & \underline{94.0} & \textbf{512.2} \\
\midrule
\multirow{9}{*}{80\%}
& SCAN (ECCV’18) & 5.1 & 18.1 & 27.3 & 3.9 & 13.1 & 19.1 & 86.6 & 22.2 & 51.9 & 67.5 & 13.8 & 41.1 & 58.6 & 255.1 \\
& SGR (AAAI’21) & 13.7 & 35.1 & 47.6 & 12.1 & 30.9 & 41.9 & 181.3 & 26.7 & 60.7 & 75.6 & 25.3 & 58.2 & 72.6 & 319.1 \\
& NCR* (NIPS’21) & 1.4 & 7.1 & 11.7 & 1.5 & 5.4 & 9.3 & 36.4 & 21.6 & 52.6 & 67.6 & 15.1 & 38.1 & 49.8 & 244.8 \\
& DECL* (ACMMM’22) & 53.4 & 78.8 & 86.9 & 37.6 & 63.8 & 73.9 & 394.4 & 64.8 & 90.5 & 96.2 & 53.7 & 81.7 & 93.0 & 473.0 \\
& MSCN* (CVPR’23) & 1.0 & 4.4 & 9.1 & 0.4 & 1.4 & 2.5 & 18.8 & 66.8 & 91.6 & 96.2 & 52.7 & 83.0 & 90.9 & 481.2 \\
& BiCro* (CVPR’23) & 3.6 & 13.9 & 20.5 & 1.7 & 7.5 & 13.0 & 60.2 & 62.2 & 88.6 & 96.1 & 47.4 & 79.2 & 88.5 & 460.5 \\
& RCL* (TPAMI’23) & 51.7 & 75.8 & 84.4 & 34.5 & 61.2 & 70.7 & 378.3 & 67.4 & 90.8 & 96.0 & 50.6 & 81.0 & 90.1 & 475.9 \\
& CREAM* (TIP’24) & \underline{58.7} & \underline{83.3} & \underline{90.1} & \underline{40.8} & \underline{67.1} & \underline{76.3} & \underline{416.3} & 68.6 & \underline{92.0} & \underline{96.4} & 52.4 & \textbf{84.8} & \underline{92.8} & \underline{487.2} \\
& L2RM* (CVPR’24) & 55.7 & 80.8 & 87.8 & 39.4 & 65.4 & 74.9 & 404.0 & \underline{69.0} & \underline{92.0} & 96.2 & \underline{52.7} & 82.4 & 92.3 & 482.6 \\
& \textbf{PCSR*} & \textbf{63.1} & \textbf{87.1} & \textbf{93.2} & \textbf{44.6} & \textbf{70.9} & \textbf{78.6} & \textbf{437.5} & \textbf{71.6} & \textbf{92.9} & \textbf{96.7} & \textbf{55.3} & \underline{83.9} &  \textbf{93.3} & \textbf{493.7} \\

\bottomrule
\end{tabular}
\caption{Image-Text Retrieval on Flickr30K and MS-COCO 1K datasets under different noise ratios. * indicates the noise robust method. The best and sub-optimal indicators are represented in \textbf{bold} and \underline{underline} respectively.}
\label{tab:noise_retrieval}
\end{table*}
To fully exploit the optimization potential of different types of pairs, we propose a training strategy named \emph{Adaptive Pair Optimization} (APO). It progressively incorporates clean, refinable, and ambiguous pairs into the training process with tailored supervision objectives.

\paragraph{Stage 1: Clean-only Training.}
Training starts with only clean pairs $\mathcal{C}$ to ensure stable initial convergence. The objective function for this phase incorporates standard triplet loss that encourages semantic alignment, along with cross-entropy loss for pseudo-label supervision and entropy regularization for promoting confident predictions. Specifically, the loss for clean data training is defined as:

\begin{equation}
\mathcal{L}_{\text{clean}} = \lambda_t\cdot\mathcal{L}_{\text{Triplet}}(I^c, T^c) + \lambda_{ce}\cdot\mathcal{L}_{\text{CE}} + \lambda_{\text{en}} \cdot \mathcal{L}_{\text{En}}
\end{equation}
where the triplet loss $\mathcal{L}_{Triplet}(I,T)$ is computed as:

\begin{align}
\mathcal{L}_{Triplet}(I,T) = \sum_{(I, T) \in \mathcal{D}} 
&\left[\hat\alpha - S(I, T) + S(I, \hat{T})\right]^+ \notag \\
+\, &\left[\hat\alpha - S(I, T) + S(\hat{I}, T)\right]^+
\end{align}
$\hat\alpha$ is defined based on the work of \cite{duan2024pc2}:

\begin{equation}
\hat\alpha = \frac{m^{{S_p}(p_i^{n}, p_j^{c})} - 1}{m - 1} \alpha,
\end{equation}
where $m$ and $\alpha$ are predefined parameters, $S_p$ is a similarity function used to calculate the similarity of two distributions.

\paragraph{Stage 2: Clean + Refinable.}
In this stage, we incorporate refinable pairs $\mathcal{R}$ into training. These samples show consistent pseudo-labels, suggesting that although their paired captions may be misaligned, their images still provide useful training signals. To correct this, we apply a \textit{caption refinement}: for each image $I_i^r \in \mathcal{R}$, we retrieve the most semantically similar caption $T_j^c$ from clean texts $T^c$ based on pseudo-label similarity.
\begin{equation}
j = \arg \max_b S_p(p_i^r, p_b^c)
\end{equation}
Based on the rematched pairs $(I_i^r, T_j^c)$, the final training loss used for refinable data training is formulated as:
\begin{equation}
\mathcal{L}_{\text{refinable}} = \mathcal{L}_{\text{Triplet}}(I_i^r, T_j^c)
\label{refinable loss}
\end{equation}

\paragraph{Stage 3: Clean + Refinable + Ambiguous.}
In the final training stage, we incorporate ambiguous pairs $\mathcal{A}$ into the optimization process to further exploit potentially informative supervision. However, due to their inherent uncertainty, we adopt a noise-robust loss function to stabilize training. Specifically, we utilize the \emph{Generalized Cross-Entropy loss} (GCE), which combines the benefits of both mean absolute error and standard cross-entropy by controlling the influence of low-confidence predictions. The GCE loss is defined as:

\begin{equation}
\mathcal{L}_{\mathrm{GCE}} = \frac{1}{B} \sum_{i=1}^{B} \left( 
\frac{1 - \big(p_i^a(\hat{q}_i^a)\big)^\gamma}{\gamma} 
+ \frac{1 - \big(q_i^a(\hat{p}_i^a)\big)^\gamma}{\gamma} 
\right)
\end{equation}
$\gamma \in (0, 1]$ is a hyperparameter controlling the robustness of the loss, set to 0.7 as recommended in \cite{zhang2018generalized}. The total loss used for ambiguous data training for this stage is formulated as:

\begin{equation}
\mathcal{L}_{\text{ambiguous}} = \lambda_t\cdot\mathcal{L}_{\text{Triplet}}(I^a, T^a) + \lambda_{gce}\cdot\mathcal{L}_{\text{GCE}} + \lambda_{\text{en}} \cdot \mathcal{L}_{\text{En}}
\label{ambiguous loss}
\end{equation}

\section{Experiments}

\subsection*{Datasets}
We evaluate our proposed method on three widely used multimodal datasets:

\textbf{Flickr30K.} This dataset comprises 31,000 images sourced from Flickr, each paired with five descriptive captions that provide detailed semantic annotations. The dataset encompasses a wide range of scenes, objects, and human activities, making it a standard benchmark for cross-modal retrieval. Following common practice, we utilize 29,000 images for training, 1,000 for validation, and 1,000 for testing.

\textbf{MSCOCO.} The MSCOCO dataset contains 123,287 images, each annotated with five captions describing various visual scenes and objects in daily life. It serves as a key benchmark for evaluating multimodal learning methods. In our experiments, we adopt the data split: 113,287 images for training, and 5,000 each for validation and testing.

\textbf{Conceptual Captions.} This large-scale dataset, curated from the web, consists of approximately 3.3 million image-text pairs. Each image is associated with an automatically mined caption, leading to a noise rate of 3\%–20\%. Following prior works \cite{huang2021learning,duan2024pc2}, we adopt the CC152K subset, which includes 150,000 training images and 1,000 images each for validation and testing.

\subsection*{Evaluation Metrics}
We adopt the standard Recall@K metrics (K = 1, 5, 10), denoted as R@1, R@5, and R@10, to evaluate image-text matching performance. These metrics measure the proportion of ground-truth instances successfully retrieved within the top-K ranked results, reflecting the effectiveness of the retrieval model in ranking relevant pairs at higher positions.

\subsection{Implementation Details}
For fair comparison, we use SGR as our cross-modal retrieval backbone, same as NCR\cite{huang2021learning}. We set the number of pseudo-label categories to $K=256$; the corresponding ablation results are presented in Table 3. For other training details, we set the batch size $B=128$, parameters $k=0.2$, $\lambda_{max}=0.9$, $\lambda_{min}=0.4$, $\beta=0.7$. Following previous studies \cite{duan2024pc2}, we set the loss weights $\lambda_{en}=10$, $\lambda_t=\lambda_{ce}=\lambda_{gce}=1$ and parameter $m=10, \alpha=0.2$. Specifically, we warm up the model for 5 epochs on Flickr30K and MSCOCO datasets, 10 epochs on CC152K dataset. After warming up, we train our model for 50 epochs, Stage 1 corresponds to epochs 1–25, Stage 2 to epochs 20–40, and Stage 3 to epochs 41–50. 

\begin{table}[t]
\centering
\renewcommand{\arraystretch}{1.2}
\setlength{\tabcolsep}{5pt}
\resizebox{1\columnwidth}{!}{
\begin{tabular}{l|ccc|ccc|c}
\toprule
% \multicolumn{8}{c}{\textbf{CC152K}} \\
% \midrule
\multirow{2}{*}{Methods} & \multicolumn{3}{c|}{Image-Text} & \multicolumn{3}{c|}{Text-Image} & \multirow{2}{*}{Rsum} \\
 & R@1 & R@5 & R@10 & R@1 & R@5 & R@10 & \\
\midrule
SCAN         & 30.5 & 55.3 & 65.3 & 26.9 & 53.0 & 64.7 & 295.7 \\
SGR          & 11.3 & 29.7 & 39.6 & 13.1 & 30.1 & 41.3 & 165.4 \\
NCR*         & 39.5 & 64.5 & 73.5 & 40.3 & 64.5 & 73.2 & 355.5 \\
DECL*        & 39.0 & 66.1 & 75.5 & 40.7 & 66.3 & 76.7 & 364.3 \\
MSCN*        & 40.1 & 65.7 & 76.0 & 40.6 & 67.1 & 76.4 & 366.7 \\
BiCro*       & 40.8 & 67.2 & 76.4 & 42.1 & 67.6 & 76.4 & 370.2 \\
RCL*         & 41.7 & 66.0 & 73.6 & 41.6 & 66.4 & 75.1 & 364.4 \\
CREAM* & 40.3 & \textbf{68.5} & \underline{77.1} & 40.2 & \textbf{68.2} & \textbf{78.3} & 373.1 \\
L2RM* & \underline{43.0} & 67.5 & 75.7 & \underline{42.8} & \underline{68.0} & \underline{77.2} & \underline{374.2} \\
\textbf{PCSR(Ours)} & \textbf{43.7} & \underline{67.7} & \textbf{77.2} & \textbf{43.1} & 67.7 & 76.3 & \textbf{375.3} \\
\bottomrule
\end{tabular}}
\caption{Image-Text Retrieval on CC152K. * indicates the noise robust method. The best and sub-optimal indicators are represented in \textbf{bold} and \underline{underline} respectively.}
\label{tab:cc152k}
\end{table}

\subsection*{Main Results}
We conduct experiments on three benchmark datasets to validate the effectiveness of our method in both synthetic and real-world noisy correspondence scenarios. To highlight the robustness of our framework, we compare it against two groups of baselines: (1) conventional cross-modal methods that do not account for noisy correspondence, including SCAN \cite{lee2018stacked} and SGR \cite{diao2021similarity}; and (2) recent noise-aware methods, including NCR \cite{huang2021learning}, DECL \cite{qin2022deep}, MSCN \cite{han2023noisy}, BiCro \cite{yang2023bicro}, RCL \cite{hu2023cross}, CREAM \cite{ma2024cross}, and L2RM \cite{han2024learning}.

\paragraph{Evaluation under Simulated Noise.} To simulate noisy training conditions, we introduce synthetic noise by randomly mismatching a portion of image-caption pairs in Flickr30K and MS-COCO datasets, with noise ratios set to 20\%, 40\%, 60\% and 80\%. As shown in Table 1, our approach consistently achieves superior performance across all noise levels. Notably, when compared with the strongest baseline CREAM\cite{ma2024cross}, our method delivers absolute improvements on Flickr30K, and MSCOCO 1K under the corresponding noise ratios. Notably, in the 80\% noise setting, our PCSR method outperforms CREAM by +4.4 R@1 on image-to-text retrieval (63.1 vs. 58.7) and +3.8 R@1 on text-to-image retrieval (44.6 vs. 40.8), demonstrating superior robustness against noisy correspondences.

\begin{table}[t]
\centering
\renewcommand{\arraystretch}{1.2}
\setlength{\tabcolsep}{4pt}
\resizebox{1\columnwidth}{!}{
\begin{tabular}{c|ccc|ccc|c}
\toprule
\multirow{2}{*}{$K$} & \multicolumn{3}{c|}{Image $\rightarrow$ Text} & \multicolumn{3}{c|}{Text $\rightarrow$ Image} & \multirow{2}{*}{Rsum} \\
 & R@1 & R@5 & R@10 & R@1 & R@5 & R@10 & \\
\midrule
32   & 69.0 & 90.0 & 95.0 & 53.1 & 78.3 & 85.9 & 471.3 \\
64  & 71.3 & 90.9 & 95.3 & 52.6 & 78.9 & 85.5 & 474.4 \\
128  & 71.4 & 91.5 & \textbf{95.9} & 52.8 & 79.2 & 85.6 & 476.4 \\
\textbf{256}  & \textbf{72.8} & \textbf{91.9} & 95.8 & \textbf{53.8} & \textbf{79.7} & \textbf{86.5} & \textbf{480.5} \\
512 & 70.0 & 91.3 & 95.4 & 53.2 & 79.1 & 86.1 & 475.1 \\
\bottomrule
\end{tabular}}
\caption{Effect of different $K$ values on retrieval performance, conducted on Flickr30k under 60\% noise.}
\label{tab:k_effect}
\end{table}

\begin{table}[t]
\centering
\renewcommand{\arraystretch}{1.2}
\setlength{\tabcolsep}{3pt}
\resizebox{1\columnwidth}{!}{
\begin{tabular}{ccc|ccc|ccc|c}
\toprule
\multicolumn{3}{c|}{Subsets} & \multicolumn{3}{c|}{Image $\rightarrow$ Text} & \multicolumn{3}{c|}{Text $\rightarrow$ Image} & Rsum \\
$S_C$ & $S_A$ & $S_R$ & R@1 & R@5 & R@10 & R@1 & R@5 & R@10 & \\
\midrule
\checkmark & \checkmark & \checkmark & \textbf{72.8} & 91.9 & \textbf{95.8} & 53.8 & \textbf{79.7} & \textbf{86.5} & \textbf{480.5} \\
\checkmark &            & \checkmark & 71.8 & \textbf{92.0} & 95.7 & \textbf{53.9} & 79.3 & 86.1 & 478.8 \\
\checkmark &\checkmark  &           & 70.9 & 91.5 & 95.9 & 52.8 & 78.7 & 85.6 & 475.4 \\
\checkmark &            &           & 70.8 & 90.6 & 94.8 & 53.1 & 78.8 & 85.5 & 473.6 \\
\bottomrule
\end{tabular}}
\caption{Component ablation study on retrieval performance, conducted on Flickr30k under 60\% noise.}
\label{tab:sample_ablation}
\end{table}

\begin{table}[t]
\centering
\renewcommand{\arraystretch}{1.2}
\setlength{\tabcolsep}{3pt}
\resizebox{1\columnwidth}{!}{
\begin{tabular}{cc|ccc|ccc|c}
\toprule
\multicolumn{2}{c|}{Components} & \multicolumn{3}{c|}{Image $\rightarrow$ Text} & \multicolumn{3}{c|}{Text $\rightarrow$ Image} & Rsum \\
CGD & APO & R@1 & R@5 & R@10 & R@1 & R@5 & R@10 & \\
\midrule
\checkmark & \checkmark & \textbf{72.8} & \textbf{91.9} & \textbf{95.8} & \textbf{53.8} & \textbf{79.7} & \textbf{86.5} & \textbf{480.5} \\
\checkmark &  $T_R$     & 71.8 & 91.7 & 95.5 & 53.4 & 78.8 & 85.6 & 476.9 \\
\checkmark &   $T_A$    & 70.8 & 90.6 & 94.8 & 53.1 & 78.8 & 85.5 & 473.6 \\
           &            & 70.8 & 90.3 & 94.4 & 53.1 & 79.0 & 85.9 & 473.5 \\

\bottomrule
\end{tabular}}

\caption{Component ablation study on retrieval performance, conducted on Flickr30k under 60\% noise.}
\label{tab:method_ablation}
\end{table}

\begin{figure}[t]
\centering
\includegraphics[width=0.9\columnwidth]{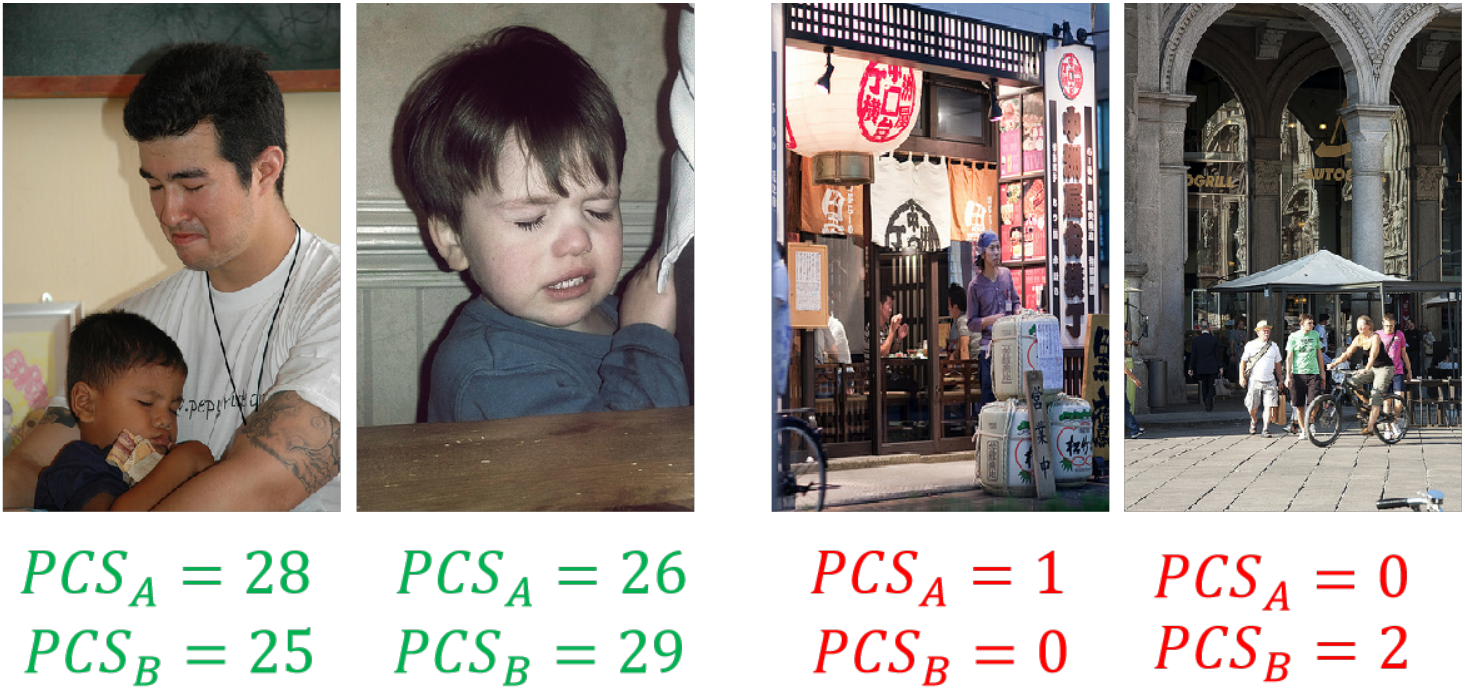} 
\caption{At epoch 40, we computed PCS values across all images using both model A and model B. These examples illustrate that semantically simpler images (left) lead to more stable pseudo-label predictions and thus higher PCS, complex semantics (right) correspond to lower PCS.}
\label{fig3}
\end{figure}

\paragraph{Evaluation under Real-World Noise.} To assess performance in realistic scenarios, we test our method on the CC152K dataset, which contains naturally occurring image-text mismatches due to imperfect web-crawled annotations. This dataset presents a more challenging setting as the noise is unlabeled and often semantically subtle. As shown in Table 2, our method demonstrates strong retrieval capabilities under real-world noise. It surpasses the best-performing baseline L2RM by a margin of 1.1 points in overall retrieval metrics. This demonstrates our framework’s ability to effectively model and utilize noisy data, maintaining robust cross-modal alignment even in complex environments.

\subsection*{Ablation Studies}

To assess the effectiveness of each PCSR component, we conduct ablation on Flickr30K with 60\% noise ratio.

\paragraph{Effect of K value.} Our PCSR framework relies on the category predictions generated by the pseudo-classifier, making the value of $K$ particularly critical. As shown in Table 3, a moderately sized $K$ allows our pseudo-classifier to model image semantics with appropriate granularity, which facilitates more accurate supervision from clean samples and enhances the quality of pseudo-caption selection. 

\paragraph{The impact of each subset.}
To evaluate the contribution of different subsets in our training strategy, we conduct an ablation study on the inclusion of clean ($S_C$), ambiguous ($S_A$), and refinable ($S_R$) pairs. As shown in Table 4, training with all three subsets achieves the best performance, with an Rsum of 480.5. Removing either ambiguous or refinable pairs leads to a noticeable drop in retrieval accuracy, confirming the complementary benefits of both pseudo-label-guided strategies. Notably, excluding $S_A$ slightly degrades performance compared to the full setting, while removing $S_R$ causes a larger drop in R@1 for both image-to-text and text-to-image. This confirms the effectiveness of exploiting confident refinable pairs via semantic realignment, as well as the robust handling of ambiguous data via the $L_{GCE}$ loss.

\paragraph{The impact of method components.}
As shown in Table 5, $T_R$ denotes the use of the refinable subset’s training method (Eq.~\eqref{refinable loss}) for both refinable and ambiguous subsets, while $T_A$ applies the ambiguous subset’s training strategy (Eq.~\eqref{ambiguous loss}) to train both refinable and ambiguous subsets. Our experiment proves that in the absence of accurate data partitioning, loss functions that were previously effective may no longer offer reliable supervisory signals.

\paragraph{Effect of PCS value.}
PCS measures the consistency of pseudo-classifier predictions, indicating an image’s semantic complexity. As shown in Figure 3, images with fewer distinctive features yield higher PCS, while semantically complex ones exhibit lower PCS due to prediction instability.

\section{Conclusion}
In this paper, we propose a novel Pseudo-label Consistency-Guided Sample Refinement (PCSR) framework for robust image-text retrieval under noisy correspondence. Our method explicitly distinguishes between clean, refinable, and ambiguous samples based on pseudo-label consistency. We design a three-stage optimization strategy: clean samples are directly trained, refinable samples are enhanced via text replacement, and ambiguous samples are optimized using noise-tolerant loss. This fine-grained division allows us to apply targeted training strategies according to sample characteristics, enabling better use of partially reliable supervision. Extensive experiments demonstrate that our method achieves substantial performance gains over existing state-of-the-art methods, especially under high-noise conditions.

\section*{Acknowledgments}
This work was partially supported by the National Science Foundation of China under Grants 62376175, U2333211 and 22494712, the 111 Project under Grant B21044, and the Natural Science Foundation of Sichuan Province under Grants 2024NSFTD0035, 2021ZDZX0011, and 2025ZNSFSC0480.

\bibliography{aaai2026}

\end{document}